\documentclass[10pt,twocolumn,letterpaper]{article}

\usepackage{iccv}
\usepackage{times}
\usepackage{epsfig}
\usepackage{graphicx}
\usepackage{amsmath}
\usepackage{amssymb}
\usepackage[dvipsnames]{xcolor}
\usepackage{mathtools}
\usepackage{multirow}

\usepackage[normalem]{ulem}

\usepackage{color,soul}
\usepackage{booktabs,makecell} % For formal tables
\usepackage{tabularx}

\usepackage[breaklinks=true,bookmarks=false]{hyperref}

\iccvfinalcopy % *** Uncomment this line for the final submission

\def\iccvPaperID{10} % *** Enter the ICCV Paper ID here

\ificcvfinal\fi
\newcommand\unfootnote[1]{%
  \begingroup
  \renewcommand\thefootnote{}\footnotetext{#1}%
  \addtocounter{footnote}{-1}%
  \endgroup
}

\begin{document}
%%%%%%%%% TITLE
\title{Multimodal Contrastive Learning with Hard Negative Sampling for Human Activity Recognition}

\author{Hyeongju Choi\textsuperscript{1}, Apoorva Beedu\textsuperscript{1}, Irfan Essa\textsuperscript{1,2}\\
\textsuperscript{1}Georgia Insitute of Technology, Atlanta, GA\\
\textsuperscript{2}Google Research, Atlanta, GA\\
{\tt\small hchoi375,abeedu3,irfan@gatech.edu}
% \and
% Apoorva Beedu\\
% Georgia Insitute of Technology\\
% Atlanta, GA\\
% {\tt\small abeedu3@gatech.edu}
% \and
% Irfan Essa\\
% Georgia Insitute of Technology\\
% Atlanta, GA\\
% {\tt\small irfan@gatech.edu}
}
\maketitle
\ificcvfinal\thispagestyle{empty}\fi

%%%%%%%%% ABSTRACT
\begin{abstract}
  Human Activity Recognition (HAR) systems have been extensively studied by the vision and ubiquitous computing communities due to their practical applications in daily life, such as smart homes, surveillance, and health monitoring. 
  Typically, this process is supervised in nature and the development of such systems requires access to large quantities of annotated data.
  However, the higher costs and challenges associated with obtaining good quality annotations have rendered the application of self-supervised methods an attractive option and contrastive learning comprises one such method.
  % As supervised training of HAR systems require large amounts of annotated data rendering it expensive and time-consuming, we extend the numerous advances in contrastive self-supervised learning (SSL) based approaches to HAR.
  % Nevertheless, a remaining challenge in SSL for HAR is the reliance on positive and negative pairs for feature representation learning. 
  However, a major component of successful contrastive learning is the selection of good positive and negative samples. 
  Although positive samples are directly obtainable, sampling good negative samples remain a challenge. 
  As human activities can be recorded by several modalities like camera and IMU sensors, we propose a hard negative sampling method for multimodal HAR with a hard negative sampling loss for skeleton and IMU data pairs. 
  We exploit hard negatives that have different labels from the anchor but are projected nearby in the latent space using an adjustable concentration parameter. 
  Through extensive experiments on two benchmark datasets: UTD-MHAD and MMAct, we demonstrate the robustness of our approach forlearning strong feature representation for HAR tasks, and on the limited data setting.
  We further show that our model outperforms all other state-of-the-art methods for UTD-MHAD dataset, and self-supervised methods for MMAct: Cross session, even when uni-modal data are used during downstream activity recognition.
\end{abstract}
\unfootnote{ICCV 2023 Workshop on PerDream: PERception, Decision making and REAsoning through Multimodal foundational modeling.} 
%%%%%%%%% BODY TEXT
\section{Introduction}

\begin{figure}[t]
    \centering
	\includegraphics[width=0.75\columnwidth]{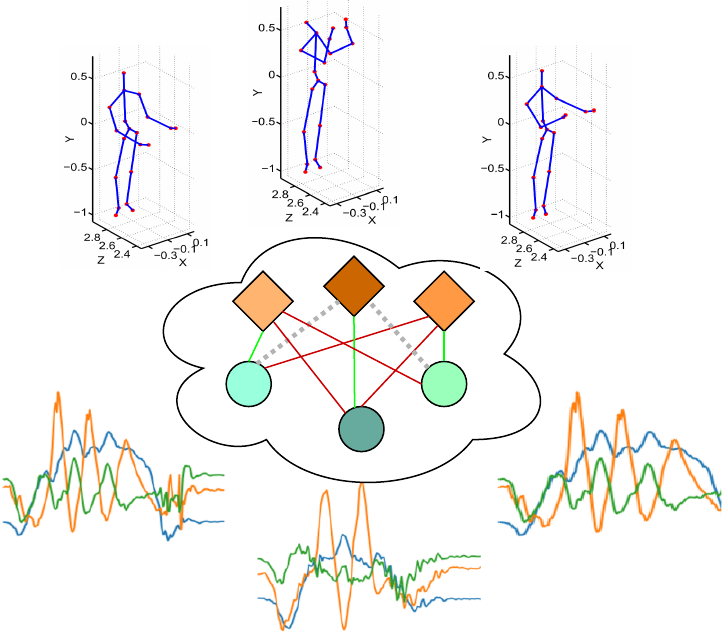}
	\caption{Illustration of \textcolor{red}{negative} and \textcolor{green}{positive} sampling methods (Best viewed in color). Uniform negative sampling would consider all $<i_n,s_k>$ samples when $n\neq k$ as negative samples, while with hard negative sampling samples that are incorrectly close to anchor(similar shades of color) is selected as negative samples while the samples very distinct(different shades) from the anchors are less preferred (marked in \textcolor{gray}{dotted gray}). }
	\label{fig:hardsamples}
\end{figure}

As large-scale datasets comprising diverse samples are increasingly helping deploy models in the real world, the need for self-supervised and unsupervised pre-training models is growing to alleviate the data annotation costs and the substantial effort needed in encoding the domain-specific knowledge.
One such family of methods, Contrastive Self-Supervised Learning (SSL), has shown great effectiveness in learning strong feature representations in many domains, including computer vision~\cite{he2020momentum,clip,Tian,tabassum2022hard,he2020momentum}, natural language processing (NLP)~\cite{gao2022simcse}, and sensor domains ~\cite{jaiswal, Bulat, haresamudram2021contrastive}. 
Nevertheless, one of the challenges of contrastive SSL is its dependence on the sampling strategy for deriving informative positive and negative pairs, and the need to train in large batches~\cite{weng2021contrastive,balestriero2023cookbook}.
Successful sampling strategies for both positive and negative pairs have been introduced and contributed to the recent progress in contrastive learning ~\cite{Tian,robinson,yang2023batchsampler,tabassum2022hard}. 
% However, negative pairs are generally sampled uniformly from the data, resulting in false negative samples that contribute to slower learning ~\cite{Boost}. 
Most constrastive methods uniformly sample the negative pairs from the data, resulting in false negative samples that contribute to slower learning, or use very large batch sizes that provide enough varies negative samples~\cite{clip,he2020momentum}.
The effect of negative samples are more pronounced in a multimodal setting where the model now needs to learn features for two different modalities. 
The effectiveness of hard negative samples (i.e., instances that are difficult to distinguish from an anchor/positive instance) has not been studied extensively in multimodal Human Activity Recognition (HAR) despite its ability to guide learning to correct its mistake more quickly. 

Several works have shown that models that use multimodal data learn stronger feature representations compared to uni-modal data setting~\cite{yuan2021multimodal,alamri2022end,nakada2023understanding,yadav2021review,subramanian2020multimodal,islam2020hamlet,choi2022multi}.
Although HAR has made significant progress in self-supervised methods \cite{tang2020exploring, khaertdinov2021contrastive,haresamudram2021contrastive} including learning from multiple devices \cite{jain2022collossl,deldari2022cocoa}, there has been comparatively less research exploring SSL methods in multimodal settings, and consequently, there has been limited exploration of sampling strategies for negative pairs in multimodal HAR systems.
Therefore, this research aims to develop a novel multimodal Human Activity Recognition (HAR) system that overcomes the challenges faced in traditional contrastive SSL methods by leveraging hard negative samples that are close to the anchor and are likely to provide the most meaningful gradient information during training as illustrated in Figure~\ref{fig:hardsamples}. 
By doing so, the proposed approach aims to be a step towards improving the foundation models for the HAR system by addressing the limitations of traditional contrastive SSL methods.

In summary, our contributions in this paper are as follows: 
\begin{itemize}
    \item We implement the hard negative sampling strategy ~\cite{robinson} into a multimodal HAR framework that mitigates false negatives and leverages hard negative samples to boost performance on feature representation learning using IMU signals and skeleton data.
    \item We perform an in-depth analysis of the effect of the adjustable concentration parameter, $\beta$, for the hard negative sampling strategy for multimodal HAR.
    \item We show the effectiveness of multimodal foundation models by using uni-modal data during during downstream task.
    \item We perform extensive experiments to evaluate our proposed method against other multimodal HAR frameworks on two publicly available multimodal datasets: UTD-MHAD~\cite{UTD-MHAD} and MMAct~\cite{mmact}.
\end{itemize}
% \emph{(1)} W

\section{Related Work}
Our work is focused on contrastive learning with hard negatives samples for multimodal HAR. In what follows, we divide and summarize the existing literature into three categories: unimodal HAR, multimodal HAR, and contrastive learning for multimodal HAR.

\subsection{Unimodal HAR}
Unimodal HAR systems, both vision based and sensor based, have been extensively studied by both communities~\cite{j8, j9, rahmani20163d,liu2017skeleton}. 
IMU-based HAR is one of the most widely used unimodal HAR approaches due to its availability on commodity platforms such as smartphones and smartwatches, and its robustness against challenges that vision-based approaches are susceptible to including occlusion, viewpoint, lighting, and background variations~\cite{sun2020human,zeng2014convolutional, chen2015deep,inoue2018deep,inoue2018deep}. 
% The most widely used approaches for IMU-based HAR use a variety of feature extractors such as CNNs, RNNs and transformers~\cite{zeng2014convolutional, chen2015deep,inoue2018deep,inoue2018deep}. 
However, IMU signals are generally noisy, and the collection of high-quality data tends to be a tedious and time-consuming endeavor.
Compared to other vision-based approaches, skeleton modality has seen a wider application in human activity recognition tasks as they directly provide body structure and pose information, is scale-invariant and robust against other challenges like variations in clothing textures and backgrounds ~\cite{sun2020human,du2015hierarchical,soo2017interpretable,shi2019skeleton,zhang2019graph}. 
% The most widely used methods for skeleton-based HAR include RNNs ~\cite{du2015hierarchical}, CNNs ~\cite{hou2016skeleton, soo2017interpretable}, and Graph Neural Networks (GNNs) ~\cite{shi2019skeleton} or Graph Convolutional Network (GCNs) ~\cite{zhang2019graph}. 
Although the unimodal-based approaches have demonstrated their effectiveness in HAR, each modality has weaknesses that other modalities can compensate for. For instance, challenges in vision-based approaches such as occlusions can be overcome by using IMU data, and the sensitivity of IMU sensors to body-worn positions ~\cite{mukhopadhyay2014wearable} can be addressed by using visual modalities such as skeleton data.

\subsection{multimodal HAR}
The advantages of aggregating information from various data modalities have been studied in computer vision for a wide range of tasks. 
For instance,~\cite{wang2019densefusion,beedu2022video} proposed a heterogeneous architecture that utilizes RGB and depth data for object pose estimation. 
In ~\cite{antol2015vqa,alamri2022end,ni2022expanding,khare2021mmbert,li2019visualbert}, Visual Question Answering (VQA) utilizes RGB images and dialog to accurately answer questions about videos. 

The success of multimodal approaches in these fields has helped address the shortcomings in the single modality based HAR approaches.
In \cite{rani2021kinematic}, Rani \etal proposed a 2D CNN-based multimodal HAR to perform human activity classification on the hand-crafted features extracted from depth images and skeleton joints. 
Franco \etal ~\cite{franco2020multimodal} proposed a skeleton and RGB-based multimodal approach where the RGB frames were used to capture the temporal evolution of actions. 
Recent multimodal HAR approaches use attention to effectively fuse features from different modalities to produce representations~\cite{islam2020hamlet,islam2022mumu,khan2021attention}. 
Although these methods have shown promising results, these methods use fully labeled data, which in itself is a non-trivial task for IMU data.

\begin{figure*}[!ht]
    \centering
	\includegraphics[width=1\textwidth]{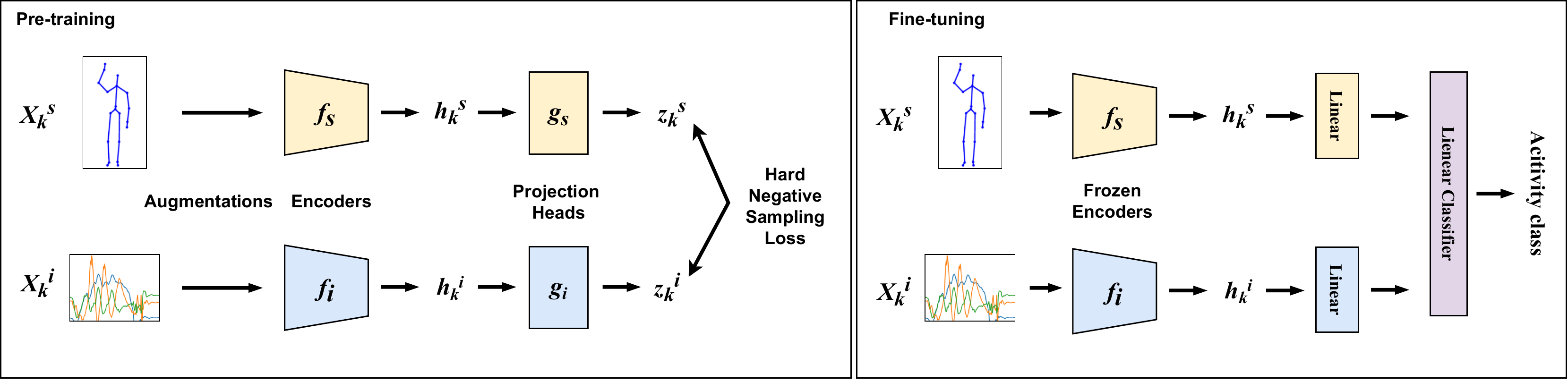}
	\caption[Model architecture]{Pre-training (left), and fine-tuning (right) training architecture for our proposed model. During the pre-training stage, the features from the skeleton data are contrasted against IMU features. During the fine-tuning stage, these features are concatenated and trained for the action classification task keeping the encoders frozen.}
	\label{fig:overview}
\end{figure*}

\subsection{Contrastive Learning for HAR}
To overcome the challenges with large annotation data, self supervised learning such as contrastive learning~\cite{chen2020simple,haresamudram2021contrastive}, or transformer based methods~\cite{haresamudram2020masked} have been widely studied in many fields including HAR for their comparable, or sometimes even better performance than the supervised learning methods. 
The main advantage of contrastive SSL approaches is that the model can be finetuned using limited labeled data when annotations are expensive to obtain. 
Since IMU data are hard to annotate and require domain-specific expertise~\cite{Zhao}, contrastive learning has been explored and applied in various unsupervised settings ~\cite{tang2020exploring, khaertdinov2021contrastive}.
There have been a few explorations of contrastive learning in the multimodal setting similar to our own~\cite{tian2020contrastive,clip,luo2022clip4clip,girdhar2023imagebind,moon2022imu2clip,he2020momentum}. 
Contrastive Multiview Coding (CMC) \cite{tian2020contrastive} maximizes mutual information between different views of the same scene, particularly between different image channels.
Alternatively, in a multimodal setting, 
Razvan \etal ~\cite{brinzea2022contrastive} proposed a contrastive SSL framework that exploits modality-specific knowledge to mitigate the problem of false negatives. 
Khaertdinov \etal ~\cite{khaertdinov2022temporal}, on the other hand, used temporal feature alignment using a dynamic time warping (DTW) in a latent space to align features in a temporal dimension. 
ImageBind~\cite{girdhar2023imagebind} use CLIP~\cite{clip} like architecture, but extend it to siz different modalities including images, text, audio, depth, thermal and IMU data. 
However, the effect of hard negative samples for multimodal HAR, to the best of our knowledge, hasn't been explored extensively. 
Inspired by~\cite{robinson} and~\cite{tian2020contrastive}, we introduce a hard negative sampling method with an adjustable hardness for multimodal HAR framework. 
This approach allows the contrastive SSL framework to exploit hard negative samples for faster training and generalize better than using randomly sampled negatives from the batch. 

\section{Methodology}
To improve the selection of negative samples in contrastive SSL for multimodal HAR, we implement a hard negative sampling method with an adjustable hardness
% in the CMC~\cite{tian2020contrastive} framework 
using two modalities: skeleton and IMU data. 
An overview of the network is shown in Figure \ref{fig:overview}.
We discuss our proposed work and hard negative sampling for contrastive learning in the sections below. 

% \vspace{-0.5em}
\subsection{Contrastive Learning for multimodal HAR}
\label{contrastiveLearning}
Contrastive learning is a method that learns to distinguish between similar and dissimilar samples. 
To this end, as a pre-training stage shown in Figure \ref{fig:overview}(left)
each sample $\{X_k^s, X_k^i\}$ representing input data for skeleton and IMU data respectively, 
% in a batch size of N 
undergoes modality-specific augmentations and is passed through modality specific encoders, $f_s, f_i$.
The resulting representations are then passed through projection heads, $g_s, g_i$, to generate projections $(z_k^s, z_k^i)$, which comprises the positive pair.
% where $z_i^s = g_s(f_s(X_{i, augmented}^s))$ and $z_i^i = g_i(f_i(X_{i, augmented}^i))$. 
The negative pairs are all the other inter-modal combinations of projections from different input instances, $(z_k^s, z_{n \neq k}^i)$ in the batch. 
Contrastive loss for the inputs can be calculated as: 
\begin{equation} \label{eq:1}
\mathcal{L} = \sum_{k=1}^{N} (l_k^{s \rightarrow i} + l_k^{i \rightarrow s})
\end{equation}
% \begin{equation} \label{eq:1}
% l_j^{i \rightarrow s} = -\log \frac{\zeta(z_i^s, z_i^i)}{\sum_{n=1}^{N} \zeta(z_i^s, z_n^i) }
% \end{equation}
% where,
% \vspace{-0.5em}
% \begin{equation} \label{eq:2}
% l_k^{i \rightarrow s} = -\log \frac{\zeta(z_k^s, z_k^i)}{\zeta(z_k^s, z_k^i) + \sum_{n=1}^{N} \zeta(z_k^s, z_n^i) }
% \end{equation}
\begin{equation*} \label{eq:2}
where, \;
l_k^{i \rightarrow s} = -\log \frac{\frac{exp(s(z_k^s, z_k^i))}{\tau}}{\sum_{n=1}^{N} \frac{exp(s(z_k^s, z_n^i))}{\tau}}
\end{equation*} and 
$s(z_k^s, z_k^i)$ is a cosine similarity function between $z_k^s$ and $z_k^i$, and $\tau$ is a temperature parameter. 

In the fine-tuning stage shown in Figure \ref{fig:overview}(right), the frozen modal-specific encoders are used to generate representations $h_k^s$ and $h_k^i$. 
These representations are then passed through a linear layer to map them into the same size, concatenated, and passed through a simple linear classification layer for the activity class prediction. 

% In our proposed framework, we modify the loss function $l_k$ in equation \ref{eq:1} with the proposed loss function described in the following section.

\subsection{Hard Negative Sampling for HAR}

Despite the success of contrastive learning in many fields, a challenge that remains is the selection of good negative samples as it has a significant impact on performance.
Inspired by ~\cite{robinson}, we introduce a hard negative sampling method for HAR to sample true negatives that have different labels from the anchor and are projected near the anchor instead of using all inter-modality pairs as negative samples as discussed in Section \ref{contrastiveLearning}.
In equations to follow, superscript ${}^+$ indicates positive samples, and ${}^-$ indicates negative samples, and $h(x)$ is the class label for the given input $x$, and $\emph{p}$ is the distribution.
The hard negative sampling method samples negatives from the distribution defined as:
\begin{equation}
 \label{eq:3}
 % \resizebox{0.91\hsize}{!}{%
\begin{gathered}
    q_\beta^- := q_\beta(x^-|h(x) \neq h(x^-)), \\
    \textnormal{where} \; q_\beta(x^-) \propto \exp^{\beta f(x)^\top {} f(x^-)}\cdot p(x^-),  \\
    h: input(x) \rightarrow labels(c)
    % }
\end{gathered}
\end{equation}
$q_B^- $ guarantees that the anchor and the negative sample correspond to different latent classes and the concentration parameter $\beta$ term up-weights the negative samples that are similar to the anchor $x$. 
The hardness term $\beta$ is a hyperparameter that can be adjusted to achieve a balance between improved learning from hard negatives and the potential harm from the approximate correction of false negatives.

The hard negative sampling objective $l_{HNL}(f)$ can be empirically obtained by adopting PU-learning (Positive unlabelled) and importance sampling to the standard contrastive learning objective, defined as:
% \begin{multline}
%         \resizebox{\columnwidth}{!}{ 
%         $\mathbb{E}_{
%             \substack{x \sim p,
%             x^+ \sim p_x^+\\
%             x^-_{{i=1}^N}\sim p^N\\
%             \{v_i\}_{i=1}^M\sim p_x^{+M}
%             }}
%         \left[-\log\frac{e^{f(x)^T f(x^+)}}{e^{f(x)^T f(x^+)}+
%         max\{\frac{1}{\tau^-}(\sum_{i=1}^N \delta e^{f(x)^T f(x^{-}_{i})} -\tau^{+}Ne^{f(x)^T f(v_i)}), Ne^{-1/t}\}} \right]
%         $}
%         \label{eq:3}
% \end{multline}
\begin{equation}
\begin{gathered}
 \resizebox{\columnwidth}{!}{ 
    $\mathbb{E}_{
    % \begin{subarray}{l}{x \sim p, x^+ \sim p_x^+}\\
    %     {\{u_i\}_{i=1}^N\sim p^N}\\
    %     {\{v_i\}_{i=1}^N\sim p_x^{+M}}
    % \end{subarray}} 
    \substack{x \sim p,\\
            x^+ \sim p_x^+\\
            x^-_{i=1:N}\sim p^N\\
            }}
    \left[-\log\frac{e^{f(x)^T f(x^+)}}{e^{f(x)^T f(x^+)} + \delta(x, x^-_{i=1:N},x^+)} \right]$
}
\end{gathered}
\label{eq:4}
\end{equation}
where $\delta(x, x^-_{i=1:N},x^+)$ is
\begin{equation*}
% \begin{gathered}
\resizebox{\columnwidth}{!}{
% \begin{aligned}
$ max\{ \frac{1}{\tau^-} (\frac{\sum_{i=1}^N e^{(\beta + 1) f(x)^Tf(x_i^-)} }{\frac{1}{N} \sum_{i=1}^N e^{\beta f(x)^Tf(x_i^-)} } -\tau^{+}Ne^{f(x)^T f(x^+)}),\;Ne^{\frac{-1}{t}}\} $
% \end{aligned}
}
\label{eq:5}
% \end{gathered}
\end{equation*}

where $\tau^+$ is the probability of anchor class, $\tau^-$ is the probability of observing a different class, $N$ is the number of negative samples and \emph{t} is the temperature.

We introduce hard negative sampling loss for HAR by simply replacing the standard contrastive loss in equation \ref{eq:1} with the hard negative sampling loss in equation \ref{eq:4}. 
Furthermore, for additional baseline comparisons, we implement debiased contrastive loss~\cite{chuang2020debiased} adapted to CMC framework: CMC-Debiased, which claims to address the issue of sampling same-label datapoints by setting $\beta=0 $ and $ \tau^+>0 $ in equation \ref{eq:4}.
In addition, we introduce hard negative sampling in SimCLR for unimodal HAR to examine the effectiveness of leveraging hard negative samples in unimodal settings.

% \vspace{-0.5em}
\subsection{Encoders}
% \subsubsection{\textbf{Unimodal Encoders}}
\label{sec:unimodalEncoder}

For the inertial encoder, we implemented CSSHAR framework ~\cite{khaertdinov2021contrastive}, a transformer-like encoder, consisting of three 1D-CNN layers with batch normalization and ReLU activation followed by positional encoding and a transformer encoder with multiple self-attention blocks to adaptively focus on the most important parts of the sensor signals ~\cite{mahmud2020human}. 
For the skeleton encoder, we implemented a hierarchical co-occurrence network introduced in ~\cite{co-occur}. 
The network takes the skeleton keypoints as input and splits the data into two unique inputs for the network: skeleton sequence and skeleton motion, i.e., the temporal difference between two consecutive frames.
% They are then fed into the network as two streams of inputs. 
% The network consists of two stages. In the first stage, the two inputs are passed through two convolutional layers followed by the transpose layer to move the joint dimension into CNN's input channels to allow the model to learn features hierarchically. 
% In the second stage, multiple convolutional layers extract global co-occurrence features from all joints, which then can be flattened and go through a couple of classifier layers for downstream tasks. 

% \vspace{-0.5em}
\emph{\textbf{Pre-training}}
\label{sec:unimodalPre}
As part of the pre-training, we apply a set of random modality-specific augmentations as suggested in ~\cite{Tian} to enhance the quality of learned embeddings. 
The inertial augmentations include \{jittering, scaling, permutation, channel shuffle\} for UTD-MHAD and \{scaling and rotation\} for MMAct. 
For skeleton data, augmentations include \{jittering, scaling, rotation, shearing, and resized crops\}. 
% More specifically, jittering is applied to every sample as a base augmentation with $p$ = 1.0, while the rest augmentations are applied to each sample with $p$ = 0.75 for randomness. 
% The hyperparameters are tuned differently depending on modality and dataset for the best results, details in Table 1.

% \vspace{-0.5em}
% \subsubsection{\textbf{multimodal Pre-training}}
% For multimodal pre-training, we use the unimodal encoders from Section \ref{unimodalEncoder} and perform identical augmentations for both modalities as in Section \ref{unimodalPre}. 
% The framework is pre-trained for 300 epochs using the Adam optimizer with a weight decay of 1e-5 and the learning rate is reduced by half when a metric has stopped improving for more than 20 epochs.
% We maintain the same training schedule as in Section \ref{unimodalPre}.
\emph{\textbf{Fine-tuning}}
In the finetuning stage, the pre-trained encoders are frozen for both unimodal and multimodal settings. 
For unimodal fine-tuning, features from the frozen encoder are flattened and passed through a linear classifier that maps the features into the number of activity classes.
For multimodal fine-tuning, features from inertial and skeleton encoders are flattened and passed through a modality-specific layer to map into a feature vector of size 256 followed by batch normalization and ReLU, then the two feature vectors are concatenated and passed through a linear classifier.

\begin{table}[t!]
\centering
\resizebox{1\columnwidth}{!}{%
% \begin{tabular}{|c|c|c|c|c|c|c|c|}
\begin{tabular}{ c c c c c c c c}
\toprule
\multicolumn{1}{c}{Method} & \multicolumn{1}{c}{Dataset} & \multicolumn{1}{c}{Modality} & \multicolumn{1}{c}{\begin{tabular}[c]{@{}c@{}}Batch \\ Size\end{tabular}} & \multicolumn{1}{c}{lr} & Temperature & \multicolumn{1}{c}{$\tau^+$} & \multicolumn{1}{c}{$\beta$} \\
% \multicolumn{1}{|c|}{Method} & \multicolumn{1}{c|}{Dataset} & \multicolumn{1}{c|}{Modality} & \multicolumn{1}{c|}{\begin{tabular}[c]{@{}c@{}}Batch \\ Size\end{tabular}} & \multicolumn{1}{c|}{lr} & Temperature & \multicolumn{1}{c|}{\tau^+} & \multicolumn{1}{c|}{\beta} \\
\midrule

SimCLR     & \multirow{4}{*}{UTD-MHAD}   & Inertial  & 128          & 0.001         & 0.5         & -  & -  \\
SimCLR-HNL & & Inertial  & 128          & 0.001         & 0.5         & 0.037     & 0.5\\
SimCLR     & & Skeleton  & 64& 0.005         & 0.5         & -  & -  \\
SimCLR-HNL & & Skeleton  & 64& 0.005         & 0.5         & 0.037     & 0.25      \\ 
% \hline
\midrule
SimCLR     & \multirow{4}{*}{\begin{tabular}[c]{@{}c@{}}MMAct:\\ Cross Subject\end{tabular}} & Inertial  & 64& 0.001         & 0.4         & -  & -  \\
SimCLR-HNL & & Inertial  & 64& 0.001         & 0.4         & 0.027     & 0.6\\
SimCLR     & & Skeleton  & 128          & 0.005         & 0.4         & -  & -  \\
SimCLR-HNL & & Skeleton  & 128          & 0.005         & 0.4         & 0.027     & 0.5\\ 
% \hline
\midrule
SimCLR     & \multirow{4}{*}{\begin{tabular}[c]{@{}c@{}}MMAct:\\ Cross Session\end{tabular}} & Inertial  & 64& 0.001         & 0.4         & -  & -  \\
SimCLR-HNL & & Inertial  & 64& 0.001         & 0.4         & 0.027     & 0.5\\
SimCLR     & & Skeleton  & 128          & 0.005         & 0.5         & -  & -  \\
SimCLR-HNL & & Skeleton  & 128          & 0.005         & 0.5         & 0.027     & 0.5\\ 
% \hline
\midrule

CMC        & \multirow{2}{*}{UTD-MHAD}   & \multirow{2}{*}{Mulitmodal} & \multirow{2}{*}{64}      & 0.001         & 0.1         & -  & -  \\
Ours(With HNL)    & &&   & 0.001         & 0.1         & 0.037     & 1.0\\
% \hline
\midrule
CMC        & \multirow{2}{*}{\begin{tabular}[c]{@{}c@{}}MMAct:\\ Cross Subject\end{tabular}} & \multirow{2}{*}{multimodal} & \multirow{2}{*}{128}     & 0.001         & 0.5         & -  & -  \\
Ours(With HNL)    & &&   & 0.001         & 0.5         & 0.027     & 1.5\\
% \hline
\midrule
CMC        & \multirow{2}{*}{\begin{tabular}[c]{@{}c@{}}MMAct:\\ Cross Session\end{tabular}} & \multirow{2}{*}{multimodal} & \multirow{2}{*}{128}     & 0.001         & 0.5         & -  & -  \\
Ours(With HNL)     & &&   & 0.001         & 0.5         & 0.027     & 0.5    \\
\bottomrule
\end{tabular}
}
\vspace{0.2cm}
\caption[Hyperparameters for pre-training]{Hyperparameters for unimodal and multimodal pre-training}
\label{tbl:sampleTbl1}
\end{table}

% \bigskip
\section{Experiments and Results}
\label{experiments}
In order to evaluate the effectiveness of the proposed approach, we conducted a set of experiments and ablations. In what follows, we discuss the setup of our training protocol and discuss the results and effects of various hyperparameters used in our training.

% \vspace{-0.5em}
\subsection{Setup}
We evaluate the performance of our proposed approaches on two benchmark multimodal datasets: UTD-MHAD ~\cite{UTD-MHAD} and MMAct ~\cite{mmact}. 
For both these datasets, we use the IMU and skeleton data for training.
% We follow the same testing protocol as in ~\cite{brinzea2022contrastive}.
Consistent with the UTD-MHAD protocol, we employed odd-numbered subjects for training and even-numbered subjects for testing purposes across all models. 
For the cross-subject evaluation in MMAct, we employed samples from the first 16 subjects for training and the remaining subjects for testing. 
For cross-session, we selected samples from the top 80\% of sessions, arranged in ascending order based on session ID, for each subject. 
We report the test accuracies for UTD-MHAD, and test accuracies and F1 scores for MMAct dataset.

For both, unimodal and multimodal training, the model is trained for 150 epochs using Adam optimizer~\cite{kingma2014adam} and a scheduler that reduces the learning rate by half when a metric has stopped improving for more than 20 epochs. 
In the fine-tuning stage, both unimodal and multimodal settings are trained for 100 epochs with the Adam optimizer.
In addition, for fair comparison with baselines, we implemented and retrained SimCLR~\cite{Tian}, a unimodal SSL framework, for unimodal encoders and also adapted CMC~\cite{tian2020contrastive} for IMU and Skeleton modalities. 
We performed 3 runs for every method, and report the average scores for all metrics.
More details on the hyperparameters for the training can be found in Table \ref{tbl:sampleTbl1}.

\begin{table}[t!]
\centering
\resizebox{1\columnwidth}{!}{%
% \begin{tabular}{|c|c|c|c|c|c|c|}
\begin{tabular}{c c c c c c c }
% \hline
\toprule
&   &     & \multicolumn{2}{c}{\multirow{2}{*}{\begin{tabular}[c]{@{}c@{}}MMAct: \\ Cross Subject\end{tabular}}} & \multicolumn{2}{c}{\multirow{2}{*}{\begin{tabular}[c]{@{}c@{}}MMAct: \\ Cross Session\end{tabular}}} \\
&   & UTD-MHAD       & \multicolumn{2}{c}{} & \multicolumn{2}{c}{} \\ 
% \hline
% &   &     & \multicolumn{2}{c|}{\multirow{2}{*}{\begin{tabular}[c]{@{}c@{}}MMAct: \\ Cross Subject\end{tabular}}} & \multicolumn{2}{c|}{\multirow{2}{*}{\begin{tabular}[c]{@{}c@{}}MMAct: \\ Cross Session\end{tabular}}} \\
% &   & UTD-MHAD       & \multicolumn{2}{c|}{} & \multicolumn{2}{c|}{} \\ \hline

Approach   & Modality & Accuracy & F-1 & Accuracy& F-1 & Accuracy\\ 
% \hline
\midrule
Supervised & Inertial & 74.65 & 62.16 & 64.24 & 80.18 & 79.17   \\
SimCLR    & Inertial & 66.74 & 52.73 & 55.29 & 70.59 & 73.34   \\
SimCLR-HNL & Inertial & 71.55 & 54.59 & 57.71 & 71.75 & 74.03   \\ 
% \hline
\midrule
Supervised & Skeleton & 93.10 & 79.76 & 80.65 & 84.36 & 84.18   \\
SimCLR     & Skeleton & 95.50 & 74.08 & 75.19 & 79.04 & 81.22   \\
SimCLR-HNL & Skeleton & \uline{95.97} & 75.15 & 76.32 & 80.06    & 81.90   \\ 
% \hline 
\midrule
% \hline
Supervised & multimodal & 94.96 &  81.37& \uline{83.23}  & \uline{89.04}& \uline{92.04}  \\
CMC-TFA* Supervised~\cite{khaertdinov2022temporal}   & multimodal & - & \textbf{84.05} & - & - & - \\
CMC        & multimodal & 95.35   & 80.78    & 83.29   & 88.91    & 91.84   \\
CMC-CMKM~\cite{brinzea2022contrastive}   & multimodal & 94.96 & 80.72 & 82.88   & - & - \\
CMC-Debiased~\cite{chuang2020debiased}   & multimodal & 95.12 & 80.80 & 83.03   & 87.93 & 91.17    \\
CMC-TFA*~\cite{khaertdinov2022temporal}   & multimodal & - & \uline{ 83.36} & - & - & - \\
Ours    & multimodal & \textbf{96.20} & {81.64}& \textbf{83.61} & \textbf{89.16}& \textbf{92.06} 
\vspace{0.2em}\\
% \hline
\bottomrule
\end{tabular}%
}
\vspace{0.2cm}
\caption[Result of feature representation learning]{Activity classification results using multimodality during pre-training and fine-tuning. The best results are in bold and the 2nd best results are underlined. * indicates that the results are reported as is from the paper, and were not reproduced.}
\label{table:results}
\end{table}

\begin{table}[t!]
\centering
\resizebox{1\columnwidth}{!}{%
% \begin{tabular}{|c|c|c|c|c|c|c|}
\begin{tabular}{c c c c c c c}
% \hline
\toprule
&   &     & \multicolumn{2}{c}{\multirow{2}{*}{\begin{tabular}[c]{@{}c@{}}MMAct: \\ Cross Subject\end{tabular}}} & \multicolumn{2}{c}{\multirow{2}{*}{\begin{tabular}[c]{@{}c@{}}MMAct: \\ Cross Session\end{tabular}}} \\
&   & UTD-MHAD & \multicolumn{2}{c}{} & \multicolumn{2}{c}{} \\ 
% \hline
% &   &     & \multicolumn{2}{c|}{\multirow{2}{*}{\begin{tabular}[c]{@{}c@{}}MMAct: \\ Cross Subject\end{tabular}}} & \multicolumn{2}{c|}{\multirow{2}{*}{\begin{tabular}[c]{@{}c@{}}MMAct: \\ Cross Session\end{tabular}}} \\
% &   & UTD-MHAD & \multicolumn{2}{c|}{} & \multicolumn{2}{c|}{} \\ \hline

Approach   & Modality & Accuracy & F-1 & Accuracy& F-1 & Accuracy\\ 
% \hline
\midrule
Supervised & Inertial & \uline{74.65} & \textbf{62.16}    & \textbf{64.24}   & \textbf{80.18}    & \textbf{79.17}   \\
SimCLR    & Inertial & 66.74 & 52.73    & 55.29   & 70.59    & 73.34   \\
SimCLR-HNL & Inertial & 71.55 & \uline{54.59}    & \uline{57.71}   & 71.75    & 74.03   \\ 
% \hline
\midrule

CMC        & Multi(Pre) & 69.77 & 50.46    & 52.62 & 71.53    & 67.79    \\
CMC-CMKM~\cite{brinzea2022contrastive}   & Multi(Pre) & 74.42   & 50.54 & 53.67 & 69.14 & 67.77   \\
CMC-Debiased~\cite{chuang2020debiased}   & Multi(Pre) & 70.70 & 49.79 & 52.08  & 68.68 & 71.95    \\
Ours    & Multi(Pre) & \textbf{75.73} & { 54.18} & { 55.74} & \uline{ 71.85}& \uline{ 74.23}
\vspace{0.1em} \\
% \hline
\bottomrule
\end{tabular}%
}
\vspace{0.2cm}
\caption[Result of feature representation learning 2]{Activity classification results using multimodality during pre-training, and using IMU data only during fine-tuning. 
The best results are in bold and the 2nd best results are underlined.}
\label{table:results2}
\end{table}

\subsection{Feature Representation Learning for HAR}

In this experiment, we compared our approach to single-modality frameworks, and other state-of-the-art multimodality frameworks.
Our primary focus in this paper is to explore the effectiveness of CMC with hard negatives for multimodal HAR, but we also investigated the effectiveness of leveraging hard negative samples in unimodal HAR by implementing the hard negative sampling loss in the SimCLR framework (i.e., SimCLR-HNL). 
Compared to SimCLR, SimCLR-HNL achieved better results in all unimodal settings for both datasets, with a performance boost ranging from 0.5\% to 4.8\%. 
For the multimodal setting, we compare the results of the supervised model using the same encoders as described in Section \ref{sec:unimodalPre}, our own implementation of CMC-CMKM~\cite{brinzea2022contrastive} using the same hyperparameters as specified in the paper, and our proposed model. 
As shown in Table \ref{table:results}, our proposed method outperformed or performed comparatively to all multimodal approaches, including supervised, in every test setting with a performance boost ranging from 0.3\% to 0.9\% compared to CMC. 
Overall, we also see that the multimodal HAR models performed better than the unimodal HAR models. 
While CMC-TFA in the supervised and self-supervised setting outperform our method, it is worth noting that their model is a stronger feature extractor as can be seen by comparing the supervised methods. 
We believe that using hard negative sampling on top of CMC-TFA would improve the accuracy even further.
% It is worth noting that the second best method for recognition was the SimCLR-HNL, outperforming the supervised multimodal methods suggesting that SimCLR with hard negative sampling can be powerful despite the lack of inertial data.
% However, for the UTD-MHAD skeleton modality, SimCLR-HNL outperformed all multimodal models except for CMC-HNL in terms of accuracy, suggesting that SimCLR with hard negative sampling can be powerful despite the lack of inertial modality.

In Table ~\ref{table:results2}, we compare the performances between SSL using uni-modal data, and multimodal data, and our method of using hard negative sampling with multimodal data. 
However, we fine-tune the model with only IMU data, to show the model's effectiveness in real-world wearable applications where having skeleton or video data is unfeasible.
We see that our method, which was trained with hard negative sampling and multimodal data outperforms all other methods including supervised methods for UTD-MHAD dataset, while surpassing all other self-supervised method for MMAct: Cross session and comparably with SimCLR with hard negative sampling for MMAct: Cross Subject. 
This shows that multimodal pre-traning with hard negative sampling is an effective traning strategy for foundational models for human activity recognition.

\subsection{Effect of \texorpdfstring{$\beta$}{Lg} in CMC}

\begin{figure}[t!]
    \centering
	\includegraphics[width=1\columnwidth]{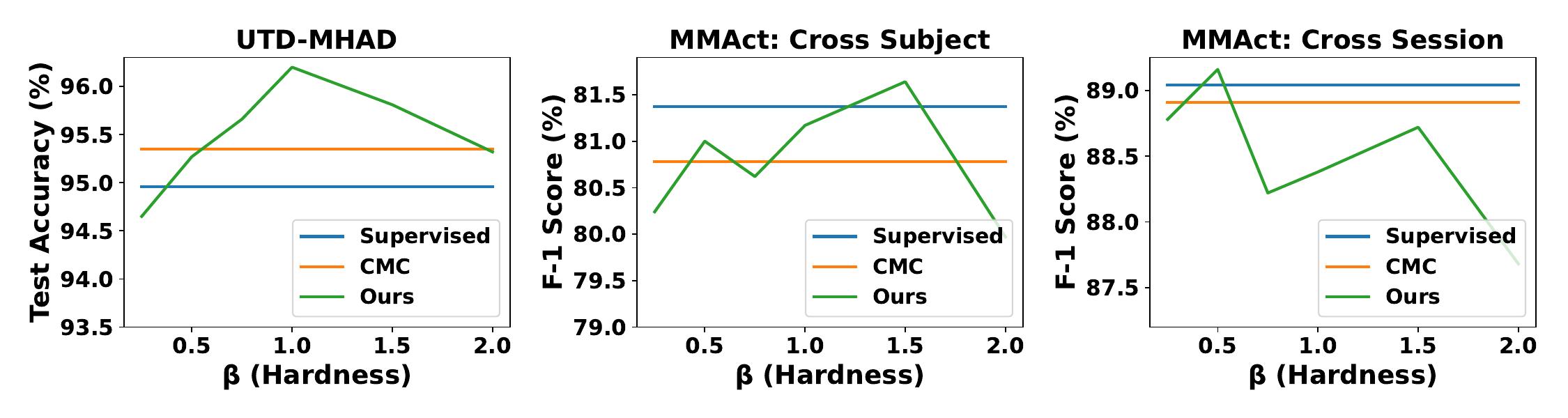}
	\caption{The effect of concentration parameter $\beta$ ranging from 0.25 to 2.0.}
	\label{fig:betaeffect}
 \vspace{-0.4cm}
\end{figure}

For the proposed approach of hard negative sampling, the concentration parameter $\beta$  determines the hardness of the negative samples and is treated as a hyperparameter. Therefore, it's crucial to tune $\beta$ adequately to balance improved learning from hard negatives and the potential negative impact of correcting false negatives. 
We adjust $\beta$ while keeping all other hyperparameters the same, using the settings specified in Table \ref{tbl:sampleTbl1} and report in Figure \ref{fig:betaeffect}. 
% which shows that the highest performance is obtained when $\beta$ falls between 0.5 and 1.5 for all three experiments. The performance difference can be up to 1.6\% depending on $\beta$. The peaks in the plots indicate an optimal range between 0.5 and 1.5, with performance dropping on either side. 
The results indicate that different datasets favor different values of $\beta$, emphasizing the importance of hyperparameter tuning to achieve optimal results even for the same dataset with different protocols.

\subsection{Fine-tuning on limited annotated data}
\begin{figure}[t!]
    \centering
	\includegraphics[width=1\columnwidth]{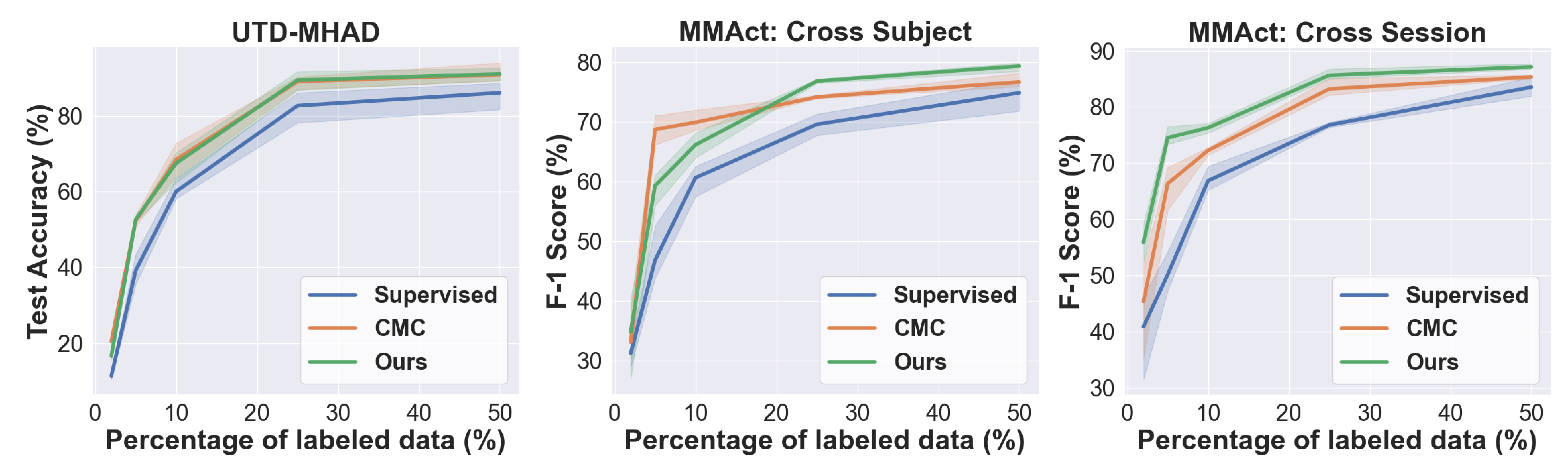}
	\caption{Performance of different multimodal models with 95\% CI for semi-supervised learning setting on UTD-MHAD and cross-subject and cross-session setting for MMAct.}
  \vspace{-1em}
	\label{fig:limitedlabels}
\end{figure}
We test our framework on the targeted setting where limited annotations are available for fine-tuning and a large unannotated dataset during pre-training. 
For each dataset, we use 100\% of unannotated samples during pre-training and limit the annotated data to 2\%, 5\%, 10\%, 25\%, and 50\% for both datasets during fine-tuning.
% We also test our approach in a semi-supervised learning setting, where we limit the number of annotated samples and compare it against the standard CMC and supervised frameworks. 
% For the SSL multimodal HAR frameworks, including CMC and CMC-HNL, we pre-train them with 100\% of unannotated samples and fine-tune the modality-specific fusion layers and linear classifier with a limited percentage of the annotated samples. 
% For the supervised HAR framework, we train the encoders and classification layers with limited annotated samples.
% Specifically, we limit the percentage of the annotated samples to 2\%, 5\%, 10\%, 25\%, and 50\% for both datasets. 
The results are reported in Figure \ref{fig:limitedlabels}. 
Across all experiments, both CMC and our framework consistently outperform the supervised model, with their performance increasing as the percentage of annotated samples decreases. 
The greatest performance boost was observed when using 5\% of the available annotated samples, with a maximum boost of up to 24\%. 
When comparing CMC and our framework, we observe no significant difference in performance for a relatively smaller UTD-MHAD dataset. 
However, for the MMAct with a cross-session protocol, our method outperforms CMC for all percentages of limited annotated labels, with a performance boost up to 10\%. 
For MMAct: cross-subject protocol, our method outperforms CMC when more than 10\% of annotated labels were available, with a performance boost of up to 2.5\%. 
Based on our findings, we conclude that our method is more effective than CMC in semi-supervised learning with limited labels.
% , particularly when more than 10\% of annotated labels are provided.
\newpage
\section{Conclusion}

In this paper, we explore the use of contrastive learning with hard negative sampling for multimodal HAR using inertial and skeleton data. 
Our goal is to mitigate the problem of false negative samples and leverage hard negative samples in multimodal HAR, which we achieve by implementing a hard negative sampling loss derived from the hardness-biased objective. 
% We propose a specific approach, CMC-HNL, which we evaluate through a series of experiments. 
Through a series of experiments, our results demonstrated that our model outperforms both supervised multimodal HAR frameworks and CMC-based multimodal HAR frameworks in various experimental settings, including limited annotated setting by learning stronger representations. 
We also explore the effect of the concentration term $\beta$ and emphasize the importance of proper tuning for optimizing the model's performance.


{\small
\bibliographystyle{ieee_fullname}
\bibliography{sample-base}
}

\end{document}